\pdfoutput=1

\documentclass[11pt]{article}

\usepackage{emnlp2021}

\usepackage{times}
\usepackage{latexsym}

\usepackage[T1]{fontenc}

\usepackage[utf8]{inputenc}

\usepackage{microtype}

\usepackage{amsmath}
\usepackage{graphicx}
\usepackage{amssymb}
\usepackage{multirow}
\usepackage{booktabs}
\usepackage{diagbox}
\usepackage{bbding}
\usepackage{enumitem}

\usepackage{bm}

\usepackage[hang,flushmargin]{footmisc}

\makeatletter
  \newcommand\smallernormal{\@setfontsize\smallernormal{9.6pt}{9.6}}
\makeatother
\makeatletter
  \newcommand\smallernormalAppendix{\@setfontsize\smallernormalAppendix{11pt}{10}}
\makeatother

\title{Extracting Event Temporal Relations via Hyperbolic Geometry}

\author{Xingwei Tan$^1$, Gabriele Pergola$^1$, Yulan He$^{1~2}$ \\
  $^1$Department of Computer Science, University of Warwick, UK\\
  $^2$Alan Turing Institute, UK\\
  \texttt{\{Xingwei.Tan, Gabriele.Pergola, Yulan.He\}@warwick.ac.uk}\\}

\date{}

\begin{document}
\maketitle
\begin{abstract}
Detecting events and their evolution through time is a crucial task in natural language understanding. Recent neural approaches to event temporal relation extraction typically map events to embeddings in the Euclidean space and train a classifier to detect temporal relations between event pairs. However, embeddings in the Euclidean space cannot capture richer asymmetric relations such as event temporal relations. We thus propose to embed events into hyperbolic spaces, which are intrinsically oriented at modeling hierarchical structures. We introduce two approaches to encode events and their temporal relations in hyperbolic spaces. One approach leverages hyperbolic embeddings to directly infer event relations through simple geometrical operations. In the second one, we devise an end-to-end architecture composed of hyperbolic neural units tailored for the temporal relation extraction task. Thorough experimental assessments on widely used datasets have shown the benefits of revisiting the tasks on a different geometrical space, resulting in state-of-the-art performance on several standard metrics. Finally, the ablation study and several qualitative analyses highlighted the rich event semantics implicitly encoded into hyperbolic spaces.\footnote{Source code is available at \url{https://github.com/Xingwei-Warwick/hyper-event-TempRel}.}
\end{abstract}

\section{Introduction}

Successful understanding of natural language depends, among other factors, on the capability to accurately detect events and their evolution through time. This has recently led to increasing interest in research for temporal relation extraction \cite{chambers2014dense, wang2020joint} with the aim of understanding events and their temporal orders. Temporal reasoning has been proven beneficial, for example, in understanding narratives \cite{cheng13}, answering questions \cite{ning20torque}, or summarizing events \cite{Wang18summ}.

\begin{figure}[!t]
\centering
\includegraphics[width=\columnwidth]{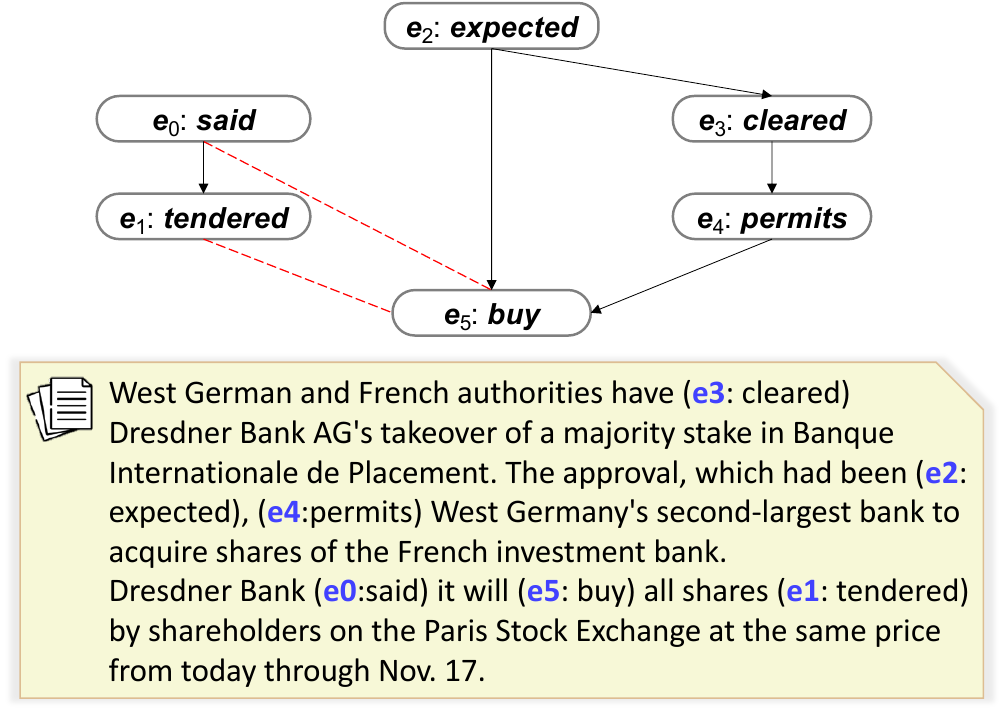}
\caption{Events annotated with temporal relations from a document excerpt. Arrow lines represent the \textit{Before} relations, while red dashed lines the \textit{Vague} ones.}
\label{fig:hier_events}
\end{figure}

However, events that occurred in text are not just simple and standalone predicates, they rather form complex and hierarchical structures with different granularity levels (Fig.~\ref{fig:hier_events}), a characteristic that still challenges existing models and restricts their performance on real-world datasets for temporal relation extraction \cite{ning2018joint,ning2018multi}.
Addressing such challenges requires models to not only recognize accurately the events and their hierarchical and chronological properties but also encode them in appropriate representations enabling effective temporal reasoning. Although this has prompted the recent development of neural architectures for automatic feature extraction \cite{ning2019improved, wang2020joint, han2019joint}, which achieves better generalization and avoids costly design of statistical methods leveraging hand-crafted features \cite{mani2006machine, chambers2007classifying, verhagen2008temporal}, the inherent complexity of temporal relations still hinders approaches that just rely on the scarce availability of annotated data.

Some of the intrinsic limitations of the mentioned approaches are due to the adopted embedding space. Existing approaches to temporal relation extraction typically operate in the Euclidean space, in which an event is represented as a point. Although Euclidean embeddings exhibit a linear algebraic structure which captures co-occurrence patterns among events, they are not able to reveal richer asymmetric relations, such as event temporal order (e.g., `\emph{event $A$ happens before event $B$}' but not vice versa). Inspired by recent works on learning non-Euclidean embeddings, such as Poincar\`{e} embeddings \cite{nickel2017poincare,tifrea2019poincar} showing superior performance in capturing asymmetrical relations of objects, we propose to learn event embeddings in hyperbolic spaces \cite{ganea2018hyperbolic}.


Hyperbolic spaces can be viewed as continuous versions of trees, thus naturally oriented to encode hierarchical and asymmetrical structures. For instance, \citet{sala18} showed that hyperbolic spaces with just two dimensions (i.e., Poincaré disk) could efficiently embed tree structures with arbitrarily low distortion \cite{Sarkar11}, while Euclidean spaces cannot achieve a comparable distortion even with an unbounded number of dimensions \cite{Linial94}.
Despite the hierarchical properties arising in modelling event relations, there are still very few studies on how to leverage those models for temporal relation extraction.
We propose two hyperbolic-based approaches with different strengths: a lightweight and efficient embedding learning method based on the Poincaré ball model, and an end-to-end deep hyperbolic neural network based on the
Riemannian optimization.

\noindent Our contributions can be summarized as follows:
\begin{itemize}[noitemsep,topsep=0pt]
    \item We propose an embedding learning approach with a novel angular loss to encode events onto hyperbolic spaces, which pairs with a simple rule-based classifier to detect event temporal relations. With only 1.5k parameters and trained in about 4 minutes, it achieves results on par with far more complex models in the recent literature.
    
    \item 
    Alternatively, we propose a hyperbolic neural network architecture for end-to-end extraction of event TempRel. 
    \item We conduct a thorough experimental assessment on MATRES and TCR, with ablation studies and qualitative analyses demonstrating the benefits of tackling the TempRel extraction task in hyperbolic spaces.
\end{itemize}

\section{Related Work}

Our work is related to at least two lines of research: one about event temporal relation extraction and another on hyperbolic neural models.


\paragraph{Event TempRel Extraction}
Approaches to TempRel extraction are largely built on neural models in recent years. 
These models have been proven capable of extracting automatically reliable event features for TempRel extraction when provided with high-quality data \cite{ning2019improved}, alleviating significantly the required human-engineer effort and yielding results outperforming the above mentioned methodologies.  In particular, \citet{ning2019improved} employed an LSTM network \cite{hochreiter1997long} to encode the textual events, taking into account their global context and feeding their representations into a multi-layer perceptron for TempRel classification. In addition, to enhance the generalization to unseen event-tuples, they simultaneously trained a Siamese network bridging common-sense knowledge across event relations. 
Similarly, \citet{han2019deep} combined a bidirectional LSTM (BiLSTM) with a structured support vector machine (SSVM), with the BiLSTM extracting the pair of events and the SSVM incorporating structural linguistic constraints across them\footnote{They compute the metrics differently and use an old version of MATRES. Thus, their results are not directly comparable}.
\citet{wang2020joint} proposed a constrained learning framework, where event pairs are encoded via a BiLSTM, enhanced with common-sense knowledge from ConceptNet \cite{Speer17} and \textsc{TemProb} \cite{NingWuPeRo18}, while enforcing a set of logical constraints at training time. The aim is to train the model to detect and extract the event relations while regularizing towards consistency on logic converted into differentiable objective functions, similarly to what was proposed in \citet{li19_logic}.

\paragraph{Hyperbolic Neural Models}
The aforementioned models are all designed to process data representations in the Euclidean space. However, several studies \cite{nickel14, Bouchard15} have shown the inherent limitations of the Euclidean space in terms of representing asymmetric relations and tree-like graphs \cite{nickel14, Bouchard15}.
%
Hyperbolic spaces, instead, are promising alternatives that have a natural hierarchical structure and can be thought of as continuous versions of trees. This makes them highly suitable and efficient to encode tree-like networks \cite{nickel2017poincare,tran2020hyperml}. 

Previous works have explored their use in embedding taxonomy for network link prediction or modeling lexical entailment. In particular, \citet{nickel2017poincare} proposed to learn word hierarchies through a negative-sampling training, based on the distance metric on the Poincaré ball.
\citet{ganea2018cones} generalized the idea of order embeddings \cite{vendrov2015order} to the Poincaré ball, levering the projected areas to infer data relations.
\citet{ganea2018hyperbolic} introduced a framework of hyperbolic neural networks composed of neural units learning and optimizing parameters in hyperbolic spaces.
Their experiments show that, without increasing the number of parameters of the models, hyperbolic neural networks outperform their Euclidean counterparts on natural language inference and detection of noisy prefixes tasks.
There have been a few attempts to revisit NLP tasks in the hyperbolic space framework by generalizing hyperbolic neural activation functions for machine translation \cite{gulcehre2018hyperbolic}, to detect hierarchical entity types \cite{lopez20}, and for document classification \cite{Zhang2020HypeHANHH}.
Compared to the above works, our model is the first attempt in devising an end-to-end hyperbolic architecture showing the benefit of addressing event TempRel extraction in hyperbolic spaces.


\section{Preliminaries}
In this section, we give a brief introduction to hyperbolic geometry and hyperbolic neural networks.

\paragraph{Hyperbolic geometry.}  
A hyperbolic space is a non-Euclidean space that has the same negative sectional curvature at every point (i.e., a constant negative curvature).
Intuitively, that a space has constant curvature implies that it keeps the same ``curveness'' at every point. An example of constant positive curvature space is a perfect globe.
On the other hand, an example of an $n$-D hyperbolic space is a hyperboloid in a $\mathbb{R}^{n+1}$ space.

One of the most widely used models of hyperbolic space was proposed by Henri Poincaré. 
The Poincaré model is an open $n$-dimensional unit ball $\mathbb{D}^n=\{x\in \mathbb{R}^n\, |\, \|x\| < 1\}$ equipped with the Riemannian metric tensor:

{\smallernormal
\begin{equation}
    g^{\mathbb{D}}_x=\lambda_x^2g^E\text{, where } \lambda_x := \frac{2}{1-\|x\|^2},
\end{equation}}
\noindent $x\in \mathbb{D}^n$, $\|\cdot\|$ denotes the Euclidean norm, and $g^E$ denotes the Euclidean metric tensor.
A two-dimensional Poincaré model is called a Poincaré disk.
A geodesic (i.e., the shortest path between two points) on the Poincaré disk is an arc, which is part of a circle perpendicular to the boundary circle.
Based on the metric tensor $g^{\mathbb{D}}_x$, the distance of two points is defined as \cite{nickel2017poincare}:

{\smallernormal
\begin{equation}
    d_{\mathbb{D}}(x,y)=\operatorname{arcosh}(1+2\frac{\|x-y\|^2}{(1-\|x\|^2)(1-\|y\|^2)}).
\end{equation}}
\noindent The Poincaré norm is the distance between the origin and the given point:

{\smallernormal
\begin{equation}
    \|x\|_{\mathbb{D}} = d_{\mathbb{D}}(0,x)=2\arctan(\|x\|).
\end{equation}}
\noindent Also, an angle $\angle ABC$ in the Poincaré model can be derived as \cite{ganea2018cones}:

{\smallernormal
\begin{equation}
\label{hyper_angle}
    \cos{(\angle(z_1,z_2))}=\frac{g^{\mathbb{D}}_x(z_1,z_2)}{\sqrt{g^{\mathbb{D}}_x(z_1,z_2)}\sqrt{g^{\mathbb{D}}_x(z_1,z_2)}},
\end{equation}}
\noindent where $z_1,z_2\in T_x\mathbb{D}^n\setminus\{0\}$ are the initial tangent vectors of the geodesics connecting $B$ with $A$, and $B$ with $C$.
An important property of the Poincaré model is its conformality with a Euclidean space, which means both of their metrics define the same angles (Eq. \ref{hyper_angle}$=\frac{\langle z_1,z_2\rangle}{\|z_1\|\|z_2\|}$, where $\langle \cdot,\cdot \rangle$ denotes Euclidean inner product).

It is known that an exponential map can be defined for each point $x\in \mathbb{D}^n$ to map any point $z\in \mathbb{R}^n(=T_x\mathbb{D}^n)$ onto $\mathbb{D}^n$ \cite{ganea2018cones}:

{\small
\begin{multline}
\label{exp_map}
    \exp_x(z) = \\ 
    \frac{\frac{1}{\|z\|}\sinh(\lambda_x\|z\|)}{1+(\lambda_x-1)\cosh(\lambda_x\|z\|)+
\lambda_x\langle x,\frac{z}{\|z\|}\rangle \sinh(\lambda_x\|z\|)}z + \\
    \frac{\lambda_x\big(\cosh(\lambda_x\|z\|)+\langle x,\frac{z}{\|z\|}\rangle \sinh(\lambda_x\|z\|)\big)}{1+(\lambda_x-1)\cosh(\lambda_x\|z\|)+
\lambda_x\langle x,\frac{z}{\|z\|}\rangle \sinh(\lambda_x\|z\|)}x.
\end{multline}}
\vspace{-12pt}

\paragraph{Hyperbolic neural networks.} In order to provide an algebraic setting for a hyperbolic space, which is not a vector space, \citet{ganea2018hyperbolic} combine the formalism of Möbius gyrovector spaces with the Riemannian geometry of the Poincaré model.
In particular, they replace the operations used in Euclidean multinominal logistic regression (MLR), Feed-Forward (FFNN) and Recurrent Neural Networks (RNN) with Möbius operations, which leads to the hyperbolic version of neural networks.
The following is the definition of the Hyperbolic Gated Recurrent Unit (HGRU):
\vspace{-8pt}

{\smallernormal
\begin{equation*}
\begin{aligned}
    z_t =& \sigma \log_0(((\textbf{W}_z\otimes h_{t-1})\oplus(\textbf{U}_z\otimes x_t))\oplus \textbf{b}_z), \\
    r_t =& \sigma \log_0(((\textbf{W}_r\otimes h_{t-1})\oplus(\textbf{U}_r\otimes x_t))\oplus \textbf{b}_r), \\
    \tilde{h}_t =& \varphi^{\otimes}((([\textbf{W}_h\operatorname{diag}(r_t)]\otimes h_{t-1})\oplus(\textbf{U}_h\otimes x_t))\oplus \textbf{b}_h), \\
    h_t =& h_{t-1}\oplus(\operatorname{diag}(z_t)\otimes((-h_{t-1})\oplus \tilde{h}_t)),
\end{aligned}
\end{equation*}
}
\noindent where $\otimes$ is the Möbius product, $\oplus$ is the Möbius addition, $\varphi^{\otimes}$ is a hyperbolic non-linearity, $\operatorname{diag}(x)$ is the square diagonal matrix of $x$, and $\{\textbf{W}_z,\textbf{U}_z,\textbf{b}_z,\textbf{W}_r,\textbf{U}_r,\textbf{b}_r,\textbf{W}_h,\textbf{U}_h,\textbf{b}_h\}$ is the parameter set.

\noindent In Hyperbolic MLR, the prediction probability of a given class $k\in {1,...,K}$ is computed as:
\vspace{-6pt}

{\smallernormal
\begin{multline}
    p(y=k|x)\propto \operatorname{exp}\big(\operatorname{sign}(\langle -\textbf{p}_k\oplus x,\textbf{a}_k\rangle)\\\sqrt{g^{\mathbb{D}}_{\textbf{p}_k}(\textbf{a}_k,\textbf{a}_k)}d_{\mathbb{D}}(x,\tilde{H}_{\textbf{a}_k,\textbf{p}_k})\big), 
\label{hyper_MLR}
\end{multline}
}

\noindent where $x\in \mathbb{D}^n$ is the output vector of the previous layer, $p_k\in \mathbb{D}^n$ and $a_k\in T_{p_k}\mathbb{D}^n\setminus\{0\}$ are parameters.

\section{Event Temporal Relation Extraction in the Hyperbolic Space }

In this section, we propose two approaches to leverage a hyperbolic space for TempRel extraction.
The first approach learns event embeddings that encode temporal order via a hyperbolic space, while the second one is an end-to-end hyperbolic neural network tailored for the TempRel extraction task.

\subsection{Hyperbolic Event Embedding Learning}
\label{sec:H-eventEmbedding}

We first explore how to learn embeddings of events in a hyperbolic space while preserving their temporal orders.
Temporal relations are asymmetric and transitive, exhibiting similar properties to hierarchical relations.
Inspired by previous successes of hyperbolic embeddings for word hierarchies \cite{nickel2017poincare,ganea2018cones}, we propose to learn event embeddings based on the Poincaré model to capture their temporal relations.

For a given text sequence containing an event pair $(u,v)$, we first extract the contextualized embeddings of the event tokens\footnote{The contextualized embeddings capture the context information around the event triggers through pre-trained language models. Thus, the resulting event embedding essentially also encodes the information about its subject and object.}, $e_u$ and $e_v$, from their, for example, ELMo \cite{peters2018deep} or RoBERTa \cite{liu2019roberta} sequence encodings, and use the exponential mapping function to map them onto a Poincaré ball. The embeddings are then further projected to a lower-dimensional space through a hyperbolic feed-forward layer:

{\smallernormal
\begin{equation}
    s_u = \mbox{HFFL}\big(\exp_0(e_u)\big) \quad s_v = \mbox{HFFL}\big(\exp_0(e_v)\big)
\vspace{-12pt}
\end{equation}}

\noindent where $\exp_0(\cdot)$ is the exponential map at the origin of a Poincaré ball as defined by Eq. (\ref{exp_map}), HFFL is a hyperbolic feed-forward layer, $s_u$ and $s_v$ are the final Poincaré embeddings of event $u$ and $v$.



To encode temporal connections in the embeddings, we want to pull events that have temporal connections close to each other, while pushing events that have no temporal relations far apart.
Thus, inspired by Poincaré embeddings \cite{nickel2017poincare}, we define the first loss term:
\vspace{-5pt}

{\smallernormal
\begin{equation}
    \label{loss_1}
    \mathcal{L}_1= \sum_{(u,v)\in \mathit{D}}{\log \frac{e^{-d_\mathbb{D}(s_u,s_v)}}{\sum_{v'\in \mathit{N(u)}}e^{-d_{\mathbb{D}}(s_u,s_{v'})}}},
\vspace{-10pt}
\end{equation}}

\noindent where $D$ is the set of event pairs that have temporal connections, $N(u)$ is the set of events that have no temporal relations with the event $u$. $s_u$ and $s_v$ denote the Poincaré embedding of event $u$ and $v$, respectively. 
For example, in the MATRES dataset, event pairs are annotated with one of the following four relations: \textsc{Before}, \textsc{After}, \textsc{Equal} and \textsc{Vague}.
We can consider the first three relations as temporal connections, and regard \textsc{Vague} as \textit{no relation} \cite{ning2019improved}.
Therefore, the set $D$ contains event pairs in the training set that have \textsc{Before}, \textsc{After}, or \textsc{Equal} labels.
The set $N(u)$ contains the events that are in the same documents with $u$ but cannot reach $u$ using only \textsc{Before}, \textsc{After}, or \textsc{Equal} edge. 
It is worth noting that because we do not encode the relation type explicitly, the order of input event pair $(u,v)$ matters.
We explicitly model the \textsc{Before} relation in the event pair $(u,v)$ (i.e., $u$ happens earlier than $v$). For event pairs with the \textsc{After} relation, we simply swap the events since \textsc{After} and \textsc{Before} are reciprocal.

\begin{figure}[!t]
\centering
\includegraphics[width=0.3\textwidth]{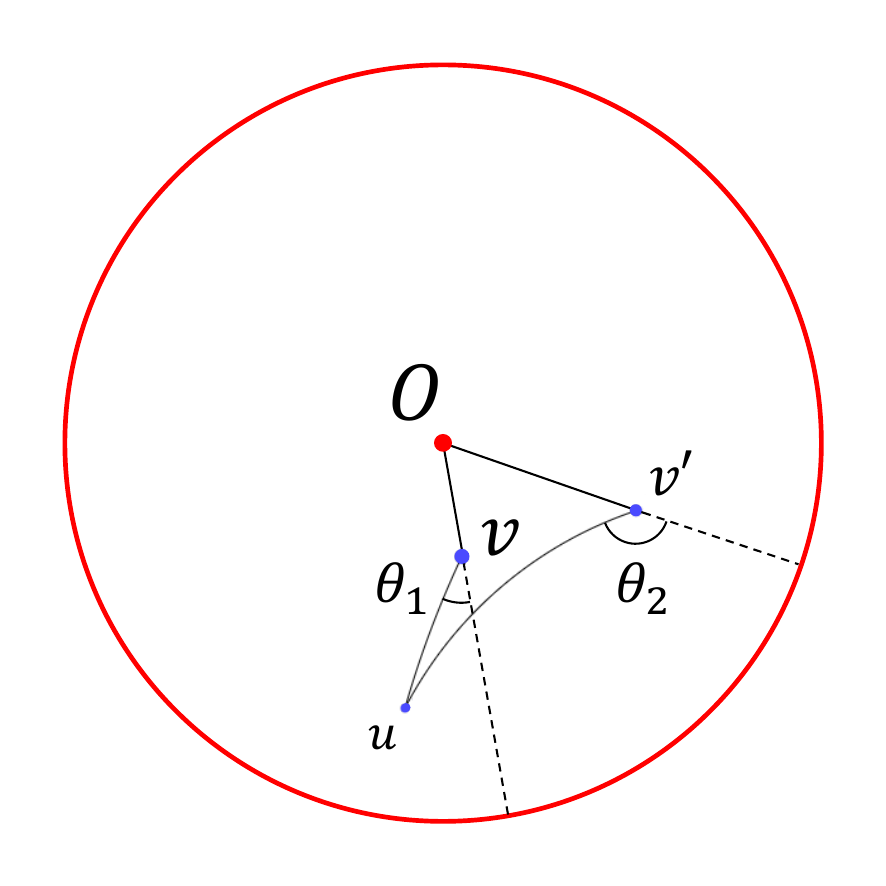}
\caption{An illustration of the Poincaré embedding used to encode two events $u$ and $v$ with known temporal relation. $\theta_1$ is the angle between the event pair $(u,v)$, while $\theta_2$ is the angle of an event pair $(u, v')$ resulting from the negative sampling process. 
}
\vspace{-10pt}
\label{fig:angular}
\end{figure}

In addition to the first loss term, we introduce a second novel loss term to enforce an angular property. 
As the example shown in Figure \ref{fig:angular}, we want to make $\angle \theta_1$ of a positive event pair $(u,v)$ smaller.
Based on preliminary tests, the angular loss can further enforce the first loss term which is driving the norm of $u$ to be larger than the norm of $v$ and thus increases the performance.
Moreover, this angular property can help to distinguish \textsc{Vague} pairs by using a threshold on the $\angle \theta_2$ to determine whether to assign the event pair to the \textsc{Vague} label.
We define the second loss term as the degree of $\angle \theta_1$, which is known to be equal to \cite{ganea2018cones}:

{\small
\begin{multline}
    \mathcal{L}_2 = \sum_{(u,v)\in \mathit{D}} \arccos \\\big(
    \frac{\langle s_u,s_v\rangle(1+\|s_v\|^2)-\|s_v\|^2(1+\|s_u\|^2)}{\|s_v\|\cdot \|s_u-s_v\|\sqrt{1+\|s_u\|^2\|s_v\|^2-2\langle s_u,s_v\rangle}}\big).
\end{multline}}


\noindent Then, we train an HFFL based on the following objective function:

{\smallernormal
\begin{equation}
    \mathcal{L}=\alpha \mathcal{L}_1 + (1-\alpha) \mathcal{L}_2,\label{eq:loss}
\end{equation}}
\noindent where $\alpha$ is a hyperparameter to balance the importance of the two loss terms.

The training objective will push $s_u$ towards the boundary while pulling $s_v$ close to the origin.
Thus, the norm of $s_u$ and $s_v$ will provide key information to determine the temporal order of an event pair, while the angle $\angle \theta_1$ can help to distinguish \textsc{Vague} event pairs.
We propose the following score function:

{\smallernormal
\begin{equation}
    \operatorname{score}(u,v)=\frac{\|s_u\|_{\mathbb{D}}-\|s_v\|_{\mathbb{D}}}{d_{\mathbb{D}}(s_u,s_v)+\epsilon}.
\end{equation}}

\noindent Based on the score function, different types of relations can be predicted using the following rules:
\vspace{-12pt}
\begin{equation}\small
    \mbox{TempRel}(u, v) =
    \begin{cases}
      \mbox{\textsc{Before}}, & \text{if}\ \operatorname{score} \in (t,1) \\
      \mbox{\textsc{After}}, & \text{if}\ \operatorname{score} \in (-1,-t) \\
      \mbox{\textsc{Equal}}, & \text{if}\ \operatorname{score} \in [-\epsilon,\epsilon] \\
      \mbox{\textsc{Vague}}, & \text{if}\ \operatorname{score} \in [-t,-\epsilon)\cup(\epsilon,t] 
    \end{cases}
    \label{eq:rules}
 \end{equation}
\noindent where the value of threshold $t\in (\epsilon,1)$ is adjusted on the validation set.


\begin{figure*}[tb]
\centering
\includegraphics[width=0.9\textwidth]{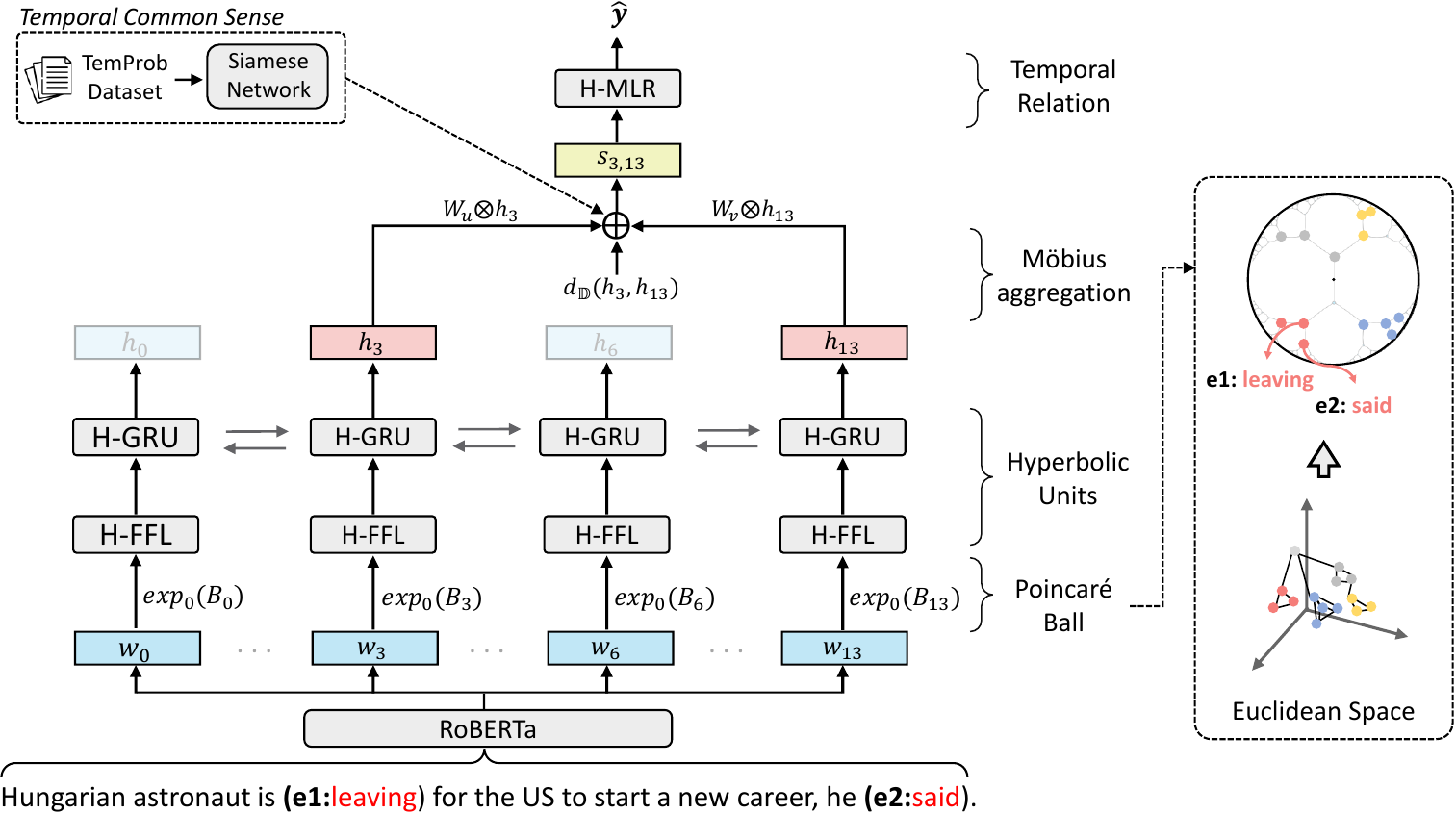}
\caption{In the hyperbolic neural architecture for temporal relation extraction, sentence tokens are first associated with standard RoBERTa vectors (within the Euclidean space). They are subsequently mapped into a Poincaré ball and processed using Hyperbolic Feed-Forward Layers (H-FFL) and Hyperbolic-GRUs (H-GRU). Then, a masking process ensures that only the event-related vectors are aggregated via Möbius operations, along with their $d_\mathbb{D}$ distance and the relevant temporal common sense, extracted by a Siamese network pre-trained on \textsc{TemProb} knowledge base. Finally, the distribution over event temporal relations is derived using a Hyperbolic Multinomial Logistic Regression (H-MLR), analogous to a traditional Softmax layer in the Euclidean space.}
\label{fig:hyper_diagram}
\end{figure*}

\subsection{Hyperbolic Neural Network for Temporal Relation Detection}
\label{sec:H-GRU}

Apart from the aforementioned approach, an alternative method is to train an end-to-end hyperbolic neural network using the cross-entropy loss.
We propose a hyperbolic neural network model for TempRel detection based on the operations defined on the Poincaré ball $\mathbb{D}^n=\{x\in \mathbb{R}^n\, |\, \|x\| < 1\}$.

Given an input sequence containing an event pair $(u,v)$, we first obtain its contextualized sentence representations from a pre-trained language model.
The representations are denoted as a matrix $B\in \mathbb{R}^{l\times d}$, where $l$ represents the sentence length and $d$ is the dimension of word embeddings. The sentence representations are then projected onto a Poincaré ball and fed into a hyperbolic feed-forward layer,
     $C = \exp_0(B)$. 
The outputs are passed to a Hyperbolic Gate Recurrent Unit (HGRU) to derive the hidden state of each word. A position masking vector $m_u$ or $m_v$, which has a value `1' in the position of an event and `0' otherwise, is applied to retrieve the hidden state representation of the corresponding event:
\begin{equation}\small
    h_u = \operatorname{HGRU}(C)\cdot m_u, \quad
    h_v = \operatorname{HGRU}(C)\cdot m_v.\nonumber
\end{equation}
The HGRU on top of the RoBERTa output can further compose the information from the event triggers and corresponding subjects and objects.

Afterward, the two event hidden states are combined by performing weighted Möbius aggregation:
\begin{equation}
\label{aggregate_func}
    s_{uv} = \textbf{W}_u\otimes h_u\oplus \textbf{W}_v \otimes h_v\oplus \textbf{b}_k.
\end{equation}
The $s_{uv}$ is further combined with the distance between the two event hidden states, $\mbox{d}_{\mathbb{D}}(h_u,h_v)$, before applying a hyperbolic non-linear function:
\begin{gather}
\label{output_layer}
    o = \varphi^{\otimes}\big(s_{uv}\oplus \mbox{d}_{\mathbb{D}}(h_u,h_v)\otimes \textbf{w}_o\big).
\end{gather}
The output $o$ is then passed to a Hyperbolic Multinomial Logistic Regression (HMLR) layer (Eq. \ref{hyper_MLR}) to generate the event temporal relation classification result, $\hat{y} = \operatorname{HMLR}(o)$.
Figure \ref{fig:hyper_diagram} shows a schematic depiction of the network architecture. 

Additionally, commonsense knowledge can be incorporated within the HGRU.
We follow \citet{ning2019improved} and use a Siamese network trained on \textsc{TemProb}\footnote{https://github.com/qiangning/TemProb-NAACL18},
discretize its output, and turn the output into categorical embeddings.
Then, we project the categorical embeddings onto a hyperbolic space and use a Riemannian optimizer to update them.
The commonsense features can be directly combined with the components of Eq. \ref{output_layer}.

\subsection{The Use of Pre-trained Language Model}

Both of our proposed methods incorporate pre-trained language models.
We investigated several ways to utilize them, and include two of them in this paper.
The first one follows \citet{ning2019improved}, which only uses the static output of the pre-trained models.
This approach is fast and allows a fair comparison with the models in \citet{ning2019improved}.
The second approach fine-tunes the pre-trained language models during training on the TempRel Extraction objective, which can achieve better performance with more cost on time.

\section{Experimental Setup}

We describe the datasets, methodologies used in the recent literature for the TempRel extraction tasks. We also briefly present the parameter setup of our experiments.
A description of the evaluation metrics can be found in the Appendix \ref{sec:metrics}.

\paragraph{Dataset}

\begin{table}[htb]
  \begin{center}
  \resizebox{\columnwidth}{!}{
  \begin{tabular}{lrrr}
  \toprule
  \bf Class & {\small MATRES Train} & {\small MATRES Test} & TCR \\ 
  \midrule 
  \textsc{Before} & $6,425$ &$427$& $1,780$\\
\textsc{After} & $4,481$ &$271$& $862$\\
\textsc{Equal} & $418$ & $30$ & $4$\\
\textsc{Vague} & $1,416$ & $109$ & $0$\\
\midrule 
Total & $12,740$ & $837$ &$2,646$ \\
  \bottomrule
  \end{tabular}}
  \end{center}
  \caption{The number of event pairs under each of the four relation classes in the MATRES and TCR datasets.}
  \label{data_table}
\end{table}

\begin{table*}[t!]
  \begin{center}
  \resizebox{\textwidth}{!}{
  \begin{tabular}{lcccccccc}
  \toprule
  ~ & \multicolumn{4}{c}{MATRES} & \multicolumn{4}{c}{TCR} \\
  \cmidrule(lr){2-5} \cmidrule(lr){6-9}
  \bf Model & P & R & Acc &  F\textsubscript{1} & P & R & Acc &  F\textsubscript{1} \\ 
  \midrule
  CogCompTime \cite{ning-etal-2018-cogcomptime}  & $61.6$ & $72.5$ & $61.6$ & $66.6$ & - & - & $68.1$ & $70.7$   \\
  LSTM* \cite{ning2019improved}& $70.2$ & $76.5$ & $67.3$ & $73.2$ & $79.6$ & $75.7$ & $75.7$ & $77.6$\\
  LSTM+knowledge* \cite{ning2019improved} & $70.2$ & $80.1$ & $70.1$ & $74.8$ & $79.3$ & $76.9$ & $76.9$ & $78.1$ \\
  LSTM+knowledge+ILP \cite{ning2019improved}  & - & - & $71.7$ & $76.7$ & - & - & $80.8$ & $78.6$  \\
  JCL-base \cite{wang2020joint} & $67.7$ & $80.3$ & - & $73.5$  & - & - & - & - \\
  JCL+multi-task+logic \cite{wang2020joint} & $72.2$ & $83.8$ & - & $77.6$ & - & - & - & -  \\
  JCL-all \cite{wang2020joint} & $73.4$ & $\bf{85.0}$ & - & $78.8$ & - & - & - & -  \\
  \midrule
  Poincaré Event Embeddings (static RoBERTa) & $72.6$ & $82.1$ & $71.9$ & $77.1$ & $81.4$ & $80.1$ & $80.1$ & $80.7$ \\
  Poincaré Event Embeddings (RoBERTa) & $74.1$ & $84.3$ & $73.7$ & $78.9$ & $85.0$ & $\bf{86.0}$ & $\bf{85.0}$ & $\bf{85.5}$ \\
   HGRU (static ELMo) & $74.4$ & $77.3$ & $69.2$ & $75.8$ & $82.9$ & $73.6$ & $73.6$ & $78.0$  \\
  HGRU (static RoBERTa) + knowledge & $73.4$ & $84.3$ & $73.4$ & $78.5$ & $83.2$ & $83.2$ & $83.2$ &  $83.2$ \\
  HGRU (RoBERTa) + knowledge & $\bf{79.2}$ & $81.7$ & $\bf{74.2}$ & $\bf{80.5}$ & $\bf{88.3}$ & $79.0$ & $79.0$ &  $83.5$ \\
  \bottomrule
  \end{tabular}}
  \end{center}
  \caption{Experimental results on MATRES and TCR. Results presented in the top half are either directly taken from the cited papers or produced by employing the original source code supplied by the authors (models denoted by `*'). Results presented in the lower half are generated from our proposed models and their variants.} 
  \label{matres_result}
\end{table*}

MATRES \cite{ning2018multi} is a TempRel dataset that is composed of news documents.
With the novel multi-axis annotation scheme, MATRES achieves much higher inter-annotator agreements (IAA) than previous temporal datasets, such as TB-Dense \cite{cassidy2014annotation}, RED \cite{o2016richer} and THYME-TimeML \cite{styler2014temporal}.
MATRES consists of documents from three sources: TimeBank (183 documents), AQUAINT (72 documents), and Platinum (20 documents).
We follow the official split in which TimeBank and AQUAINT are used for training, while Platinum is used for testing.
We further split 20\% of the training data as the validation set.

Temporal and Causal Reasoning (TCR) \cite{ning2018joint} is another dataset that adopts the annotation scheme defined in MATRES. It is a much smaller dataset, with just $25$ documents and $2.6$K TempRels.
Due to the TCR limited size, we follow \citet{ning2019improved} by using the temporal relations in TCR to test the model trained on MATRES. The statistics of the data used in our experiments is shown in Table \ref{data_table}.

\paragraph{Compared Methods}

We compare our proposed Poincaré event embedding (\textsection  \ref{sec:H-eventEmbedding}) and hyperbolic architecture for TempRel extraction (\textsection  \ref{sec:H-GRU}) with the following baselines:\\
\noindent\underline{CogCompTime} \cite{ning-etal-2018-cogcomptime} is a pipeline system based on semantic features and structured inference. \\
\noindent\underline{LSTM} denotes a TempRel detection model based on LSTMs \cite{ning2019improved}. It has two variants, one with the incorporation of the commonsense knowledge (\textit{LSTM+knowledge}) and one with the additional global inference via Integer Linear Programming (\textit{LSTM+knowledge+ILP}). \\
\noindent\underline{Joint Constrained Learning} \cite{wang2020joint} conducts joint training on both temporal and hierarchical relation extraction based on RoBERTa and Bi-LSTMs. It incorporates logic constraints and commonsense knowledge establishing the current state-of-the-art results on the MATRES dataset.


\paragraph{Parameter Setup}

Based on the results of preliminary experiments, the contextualized embeddings are produced from RoBERTa in both of our proposed approaches.
We conducted preliminary experiments to determine the best dimension for the Poincaré embeddings but found the variations in performance had no statistical significance.
Thus, we adopted the 2D Poincaré embeddings for the sake of simplicity and ease of visualization.
More details about the model architecture and hyperparameter setting can be found in the Appendix \ref{sec:hyperparameter}.


\section{Experimental Results}



\paragraph{Overall Comparison} 

Existing methodologies adopted for TempRel extraction commonly leverage several auxiliary components, such as external commonsense knowledge and multi-task objectives. Therefore, to better understand the impact made by the adoption of the hyperbolic geometry, we conduct additional experiments over ablated versions of the baseline models. In particular, in Table \ref{matres_result}, results for the \texttt{LSTM} and \texttt{LSTM$+$knowledge} are produced by employing the original code provided by the authors\footnote{\url{https://github.com/qiangning/NeuralTemporalRelation-EMNLP19}}.
Other results of compared methods are directly taken from the cited papers.
For consistency and fair comparison with \citet{ning2019improved}, we test our method with inputs from the ELMo \cite{peters2018deep}, which has been reported achieving the best overall results for the models proposed in 
\cite{ning2019improved}.

We observe that the proposed Poincaré event embedding learning method presented in Section~\ref{sec:H-eventEmbedding} outperforms \texttt{LSTM} and its variants which rely on fairly complex auxiliary features and constraints on both the MATRES and the TCR datasets, and produced more accurate event TempRel detection results even when compared to the \texttt{JCL-base} model. It is worth noticing that the \texttt{Poincaré event embeddings (static RoBERTa)} are trained with a shallow network with just $1.5$k parameters and only takes about $4$ minutes to train on a single RTX 2080 Ti. If further fine-tuning RoBERTa on MATRES, we observe a further improvement on F\textsubscript{1} by 1.8\%.

Using our proposed alternative end-to-end HGRU model, \texttt{HGRU (static ELMo)} outperforms both the standard \texttt{LSTM} model and the variant incorporating commonsense knowledge (\texttt{LSTM$+$knowledge}) on MATRES.
This verifies that the hyperbolic-based method is more efficient than its Euclidean counterparts. 
In order to fairly compare with the state-of-the-art model \cite{wang2020joint} on this task, we utilize RoBERTa and the auxiliary temporal commonsense knowledge since \citet{wang2020joint} also fine-tunes RoBERTa on MATRES and uses external commonsense knowledge.
The results show that \texttt{HGRU (RoBERTa) $+$ knowledge} outperforms \texttt{JCL-base} \cite{wang2020joint} significantly by 7\% in F\textsubscript{1}.
It even outperforms \texttt{JCL-all}, which further incorporates logic constraints and multi-task learning.

On the TCR dataset, the proposed hyperbolic-based methods also see similar improvement over existing methods.
In terms of the difference between the two proposed methods, Poincaré Events Embeddings achieve a higher F\textsubscript{1} score on TCR.
The reason is that HGRU tends to predict more \textsc{Vague} labels, but there is no \textsc{Vague} in TCR.
Interestingly, both proposed methods predict less \textsc{Vague} labels when using static RoBERTa.
A detailed breakdown of results for each temporal relation and training cost is presented in the Appendix \ref{sec:ref_breakdown}.

\paragraph{Ablation Study}

In order to study the impact of different components of HGRU on event TempRel extraction, we conduct the ablation study.

First, the HGRU layer is removed and the contextual embeddings of events are directly fed into hyperbolic FFNN (HFFNN) while all the other hyperparameters are frozen.
The resulting performance of HFFNN is significantly lower than HGRU, which indicates that the temporal information is spread across different time steps of the pre-trained language model output, and it is better encoded by a recurrent architecture.
HGRU w/o $\mbox{d}_{\mathbb{D}}$ shows the impact of the hyperbolic distance feature $\mbox{d}_{\mathbb{D}}(h_u,h_v)$ and its parameter $\textbf{w}_o$ in Eq. \ref{output_layer}.
The results show that the hyperbolic distance between the hidden states of two events encodes relevant information to predict the event temporal relations.

\begin{table}[tb]
  \begin{center}
  \resizebox{\columnwidth}{!}{
  \begin{tabular}{lcc}
  \toprule
  \bf Model & \bf Acc & F\textsubscript{1} \\ 
  \midrule
  HFFNN (static ELMo) & $67.1$ & $72.7$ \\
  HFFNN (static RoBERTa) & $67.9$ & $73.6$ \\
  HGRU (static ELMo) $\quad\,$ w/o $\mbox{d}_{\mathbb{D}}$ & $67.7$ & $74.3$ \\
  HGRU (static ELMo) &  $69.2$ & $75.8$ \\
  HGRU (static RoBERTa) w/o $\mbox{d}_{\mathbb{D}}$  & $71.0$ & $76.1$ \\
  HGRU (static RoBERTa) with EMLR & $71.8$ & $76.8$ \\
  HGRU (static RoBERTa) & $72.2$ & $77.6$ \\
  HGRU (static RoBERTa) + knowledge & $73.4$ & $78.5$ \\
  HGRU (RoBERTa) + knowledge & $\bf{74.2}$ & $\bf{80.5}$ \\
  \bottomrule
  \end{tabular}}
  \end{center}
  \caption{Ablation experiments on the MATRES dataset.} 
  \label{ablation}
\end{table}

Although \citet{ganea2018hyperbolic} reported that mixing hyperbolic neural networks with Euclidean Multinomial Logistic Regression (EMLR) can at times achieve better performance than pure hyperbolic networks, we observe no significant difference on the MATRES dataset. Finally, as discussed earlier, fine-tunning RoBERTa gives better performance compared to static RoBERTa and ELMo.

\noindent Additionally, our ablation study on the Poincaré Event Embedding shows that without the angular loss ($\alpha=1$, Eq. \ref{eq:loss}) it can only achieve $62.4$ in F\textsubscript{1}, compared to $77.1$ while using it (static RoBERTa).






\paragraph{Case Study}

\begin{figure}[tb]
\centering
\includegraphics[width=\columnwidth]{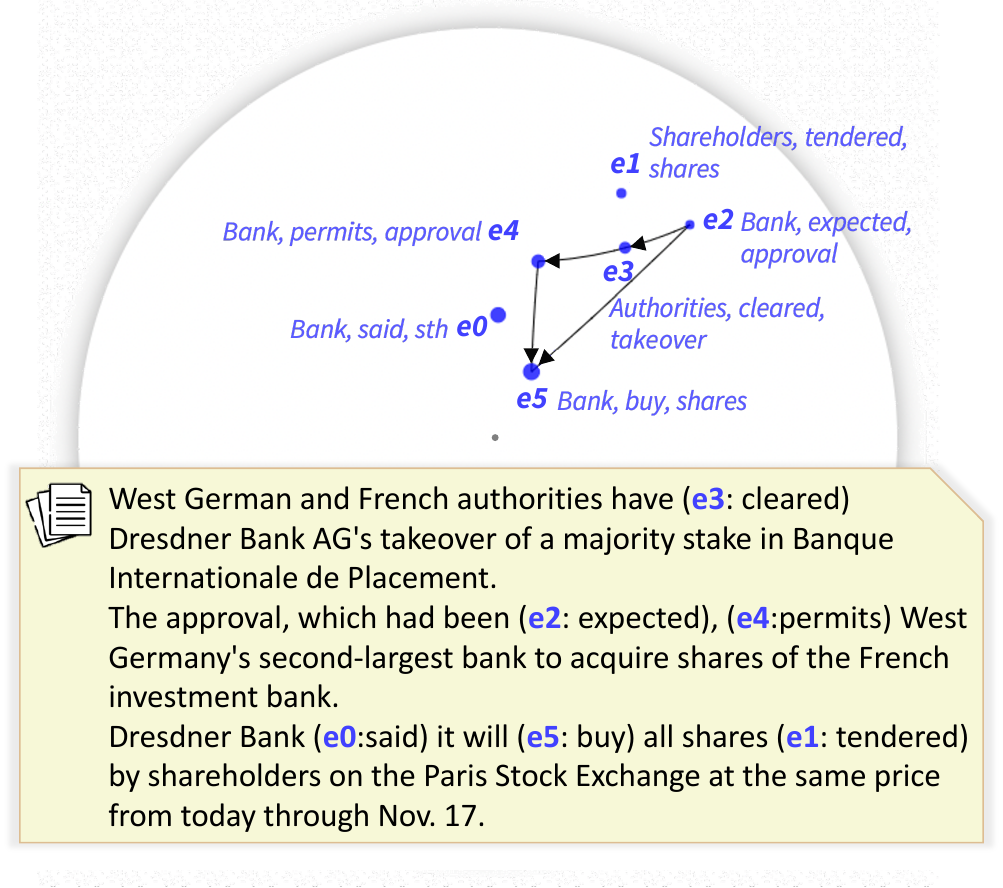}
\caption{A document excerpt from the MATRES dataset and the related temporal event embedding generated by the Poincaré embedding method.} 
\label{fig:poinc_doc}
\end{figure}

Figure \ref{fig:poinc_doc} shows a set of Poincaré event embeddings resulting from the method proposed in section \ref{sec:H-eventEmbedding}\footnote{Each point corresponds to the contextualized embedding of an event represented by its predicate, e.g., $e1$: \emph{said}, $e2$: \emph{expected}, etc. We denote each point in Figure \ref{fig:poinc_doc} by a tuple \emph{(subject, predicate, object)} for easy inspection.}. The events are numbered according to the temporal relations detected by the model (a smaller number denotes an earlier event). 
Among them, it is worth noting the two temporal paths between the \emph{expected approval from the bank} (\textbf{e2}) and the \emph{final acquisition of shares} (\textbf{e5}), i.e., $e2\rightarrow e5$ and $e2\rightarrow e3\rightarrow e4\rightarrow e5$. The first direct path is accompanied by a more fine-grained path, specifying the clearance granted from the authorities (\textbf{e3)} to permit (\textbf{e4}) the bank acquisition and the consequential buying of tendered shares (\textbf{e5)}. 
The model has encoded more recent events closer to the origin, and events in the past closer to the border, while simultaneously shaping a hierarchical structure to link their information with different granularity, for a resulting hierarchical structure with asymmetric connections.


\section{Conclusion}

In this paper, we proposed to model event temporal relations overcoming the limitations of Euclidean representations, and designing two TempRel extraction methods using hyperbolic geometry.
The first approach highlighted the convenience of learning event embedding in the Poincaré ball, achieving performance on par with recent methodologies using a simple rule-based classifier.
Then, we designed a hyperbolic neural network, incorporating temporal commonsense, outperforming state-of-the-art models on the standard datasets.
Finally, a qualitative analysis pointed out the inherent advantage of employing hyperbolic spaces to encode asymmetric relations.
In the future, we plan to extend our approaches to a wider spectrum of event relations, including causal and sub-event relations.

\section*{Acknowledgements}

This work was funded in part by the UK Engineering and Physical Sciences Research Council (grant no. EP/V048597/1, EP/T017112/1). YH is supported by a Turing AI Fellowship funded by the UK Research and Innovation (grant no. EP/V020579/1).

\bibliographystyle{acl_natbib}
\bibliography{emnlp2021}

\clearpage
\appendix
\section*{Appendix}
\renewcommand{\thesubsection}{\Alph{subsection}}
\setcounter{table}{0}
\renewcommand{\thetable}{A\arabic{table}}
\setcounter{footnote}{0}

\subsection{Model Architecture and Hyperparameter Setting}
\label{sec:hyperparameter}

The implementation of the proposed models is based on the \textit{geoopt} package \cite{geoopt2020kochurov}.
RoBERTa used in the experiments is downloaded from  Huggingface\footnote{\url{https://huggingface.co/}} \cite{wolf-etal-2020-transformers}.
ELMo is downloaded from AllenNLP\footnote{\url{https://github.com/allenai/allennlp}}.

The implementation architecture of the TempRel HGRU model is shown in Table \ref{tab:hyperparameters}. We use the RoBERTa base model, whose output dimension $d_1$ is $768$.
The dimension of the hidden states $d_2$ is set to $128$.
The commonsense embedding dimension $d_3$ is set to $32$.
The FFNN output dimension $d_4$ is $64$.
The activation function before the HMLR layer is ReLU.
We use Riemannian Adam \cite{becigneul2018riemannian} to optimize the hyperbolic parameters, and the standard Adam optimizer for parameters in the Euclidean space.
The learning rate for RoBERTa fine-tuning is $1\times 10^{-5}$, and for other parameters in our proposed models is set to $1\times 10^{-3}$.

For the Poincaré event embeddings (\textsection \ref{sec:H-eventEmbedding}), the contextualized embeddings are produced from RoBERTa-base.
We conducted preliminary experiments to determine the best dimension for the Poincaré embeddings, and found variation in performance having no statistical significance.
We then adopted the 2D embeddings for the sake of simplicity and ease of visualization.
The weight $\alpha$ for balancing two loss terms is set to $0.5$.
The number of negative samples is set to $1$.

\begin{table}[!t]
  \begin{center}\small{
  \begin{tabular}{l|ccc|ccc}
  \toprule
  ~ & \multicolumn{3}{c|}{MATRES} & \multicolumn{3}{c}{TCR} \\
  \bf Relation & P & R & F\textsubscript{1} & P & R & F\textsubscript{1} \\ 
  \midrule
  \multicolumn{7}{l}{LSTM \cite{ning2019improved}} \\
  \midrule
  Before  &$73.0$ & $85.0$ & $78.6$ & $87.8$ & $80.1$ & $83.8$   \\
  After & $66.5$ & $80.4$ & $72.8$ & $66.4$ & $75.5$ & $70.6$\\
  Equal & $0$ & $0$ & $0$ & $0$ & $0$ & $0$ \\
  Vague  & $33.3$ & $3.7$ & $6.6$ & - & - & -  \\
  \midrule
  \multicolumn{7}{l}{Poincaré Event Embeddings (ours)} \\
  \midrule
  Before  & $75.5$  & $90.2$ & $82.2$ & $90.9$ & $87.0$ & $88.9$   \\
  After & $71.8$ & $84.5$ & $77.6$ & $77.0$ & $81.2$ & $79.1$ \\
  Equal & $0$ & $0$ & $0$ & $0$ & $0$ & $0$ \\
  Vague  & $37.5$ & $3.8$ & $5.1$ & - & - & - \\
  \midrule
  \multicolumn{7}{l}{TempRel HGRU (ours)} \\
  \midrule
  Before  & $82.4$  & $85.7$ & $84.0$ & $95.3$ & $77.5$ & $85.5$   \\
  After & $74.6$ & $84.5$ & $79.2$ & $77.4$ & $82.6$ & $80.0$\\
  Equal & $0$ & $0$ & $0$ & $0$ & $0$ & $0$ \\
  Vague  & $30.2$ & $23.9$ & $26.7$ & -  & - & - \\
  \bottomrule
  \end{tabular}}
  \end{center}
  \caption{A breakdown of evaluation results by temporal relation types. \label{class_specific}}
\end{table}

\begin{table*}[!t]
\setlength{\extrarowheight}{5pt}
\resizebox{\textwidth}{!}{
    \centering
    \begin{tabular}{p{2cm}cl}
    \toprule[1pt]
\textbf{Input}&& Sentences $\{s_{i}\}_{i=1}^{N}$ each of which containing one of more event pairs $(u,v)$ with  a token \\ &&  sequence $s_{i}=\{x_{ij}\}_{j=1}^{L}$,
the masks $(m_u,m_v)$ indicate the position of the events in a sentence.\\ 


\textbf{Contextual embedding}& ELMo or RoBERTa & Encoded by a pre-trained language model $\{\boldmath{x}_{ij}\}_{j=1}^{L} - \{\rm{ELMo / RoBERTa}\}\rightarrow$: $\{\boldmath{c}_{ij}\}_{j=1}^{L} \in \mathbb{R}^{d_1\times L}$   \\


\textbf{Hyperbolic encoding}& HGRU & Map the contextual embeddings onto hyperbolic space $\{\bm{c}_{ij}\}_{j=1}^{L} - \{\rm{expmap}_0\}-\{\rm{HGRU}\}\rightarrow$: $\bm{h}_{u},\bm{h}_{v}  \in \mathbb{R}^{d_2}$    \\


\textbf{H-distance}& & Compute the distance between two points on the Poincaré model $d_\mathbb{D}(\bm{h}_u,\bm{h}_v) \in \mathbb{R}$    \\ 


\textbf{Commonsense}& VerbNet & Extract score from the pre-trained VerbNet $\bm{x}_{u},\bm{x}_{v}-\{\rm{VerbNet}\}\rightarrow$:$\bm{k} \in \mathbb{R}$, \\ 
&  & discretize the output score $\bm{k} - \{discretize\} - \{embedding\}\rightarrow$: $\bm{g}\in \mathbb{R}^{d_3}$\\


\textbf{Combine}& H-concat & Hyperbolic concatenation $\bm{h}_u \oplus \bm{h}_v \oplus d_\mathbb{D}(\bm{h}_u,\bm{h}_v) \oplus \bm{g} - \{relu\}\rightarrow $: $\bm{o} \in \mathbb{R}^{d_4}$    \\ 


\textbf{Classification}& HMLR & Hyperbolic multinomial logistic regression $\bm{o} - \{HMLR\} -\{Softmax\}\rightarrow $: $\hat{y}$    \\ 

\bottomrule[1pt]
    \end{tabular}}
    \caption{TempRel HGRU framework and hyperparameters}
    \label{tab:hyperparameters}
\end{table*}

\subsection{A Breakdown of Evaluation Results by Temporal Relation Types}
\label{sec:ref_breakdown}
Table \ref{class_specific} shows the model performance with respect to each relation type.
The two proposed methods show increased performance on \textsc{Before} and \textsc{After} relations which take the majority in the MATRES and TCR datasets.
The F\textsubscript{1} scores on \textsc{Equal} and \textsc{Vague} are  considerably low.
This is probably due to the limited number of \textsc{Equal} instances.
For the \textsc{Vague} instances, even human annotators are unable to determine the relations.
Interestingly, our TempRel HGRU with RoBERT fine-tuning has significantly higher F\textsubscript{1} score on \textsc{Vague} compared to LSTM or Poincaré event embeddings.
Our hypothesis for this paper is that hyperbolic geometry can improve the extraction of asymmetric relations, but \textsc{Vague} is symmetric.
The improvement on \textsc{Vague} is unexpected.
Further investigation is needed to find out the reason behind this phenomenon.

\subsection{Training Cost}

The training time of the proposed TempRel HGRU (static RoBERTa) is about $2$ minutes per epoch on the MATRES dataset (single Nvidia RTX 2080Ti GPU, batch\_size=250). Evaluation on the validation set takes around $23$ seconds. The best validation score tends to appear at around the 10th epoch.
The proposed Poincaré event embedding method is much more efficient.
The entire training of the Poincaré event embeddings (static RoBERTa) only takes about $4$ minutes (single Nvidia RTX 2080 Ti, 20 epochs).

Our HGRU with RoBERTa fine-tuning takes about $32$ minutes per training and validation epoch to train on a single Nvidia RTX 3090 GPU.
The best validation score also tends to appear at around the 10th epoch.

For comparison, the state-of-the-art model \cite{wang2020joint} requires about $8.5$ minutes per epoch in training if not fine-tuned the RoBERTa.
If fine-tuning RoBERTa, it takes about $32.5$ minutes per epoch for training (including validation).

\subsection{Evaluation Metrics}
\label{sec:metrics}

For the evaluation scores used in Table \ref{matres_result}, we follow the widely adopted evaluation metrics proposed in \citet{ning2019improved}, which is also adopted by \citet{wang2020joint}.
In particular, given the four temporal relation types: \textsc{Before}, \textsc{After}, \textsc{Equal} and \textsc{Vague}, 
a confusion matrix can be built with the row $C_{[k,:]},k\in \{b,a,e,v\}$ representing the gold label of class $k$; the column $C_{[:,k]}$ representing model prediction of class $k$.
The metrics are computed as follows:\\
\noindent\underline{Accuracy}. $Acc=(C_{b,b}+C_{a,a}+C_{e,e}+C_{v,v})/S$, where $S$ is the sum of all the values in the confusion matrix.\\
\noindent\underline{Precision, recall, and F\textsubscript{1}}. The \textit{precision} $P=(C_{b,b}+C_{a,a}+C_{e,e})/S_1$, where $S_1$ indicates the sum of the first three columns in the confusion matrix. The \textit{recall} $R=(C_{b,b}+C_{a,a}+C_{e,e})/S_2$, where $S_1$ denotes the first three rows in the confusion matrix. $F_1=2PR/(P+R)$.

\end{document}